%% file: main.tex
 \documentclass[acmsmall,screen]{acmart}
 \usepackage{lscape}
 \usepackage{rotating}
\usepackage{stfloats}
\usepackage{subfig}
\usepackage{svg}
\usepackage{graphicx}
\usepackage{multirow}
\usepackage{amsmath}
\AtBeginDocument{%
  \providecommand\BibTeX{{%
    \normalfont B\kern-0.5em{\scshape i\kern-0.25em b}\kern-0.8em\TeX}}}

\setcopyright{acmcopyright}
\copyrightyear{2024}
\acmYear{2024}
\acmDOI{X.X}

\acmConference[ACM Trans. Multimedia Comput. Commun. Appl. '24]{Make sure to enter the correct
  conference title from your rights confirmation email}{}{NY}
\acmPrice{15.00}
\acmISBN{978-1-4503-XXXX-X/18/06}

\begin{document}

\title{From CNNs to Transformers in Multimodal Human
Action Recognition: A Survey}

\author{Muhammad Bilal Shaikh}
\email{m.shaikh@ecu.edu.au}
\orcid{0000-0001-9042-5018}
\author{Douglas Chai}
\orcid{0000-0002-9004-7608}
\author{Syed Mohammed Shamsul Islam}
\orcid{0000-0002-3200-2903}
\email{d.chai@ecu.edu.au}
\affiliation{%
  \institution{Edith Cowan University}
  \streetaddress{270 Joondalup Drive}
  \city{Perth}
  \state{Western Australia}
  \country{Australia}
  \postcode{6027}
}

\author{Naveed Akhtar}
\orcid{0000-0003-3406-673X}
\affiliation{%
  \institution{School of Computing and Information Systems, The University of Melbourne}
  \streetaddress{700 Swanston Street}
  \city{Carlton}
  \state{Victoria}
  \country{Australia}
  \postcode{3010}
}
\email{naveed.akhtar@uwa.edu.au}

\renewcommand{\shortauthors}{Shaikh and Chai, et al.}

\input{0.abstract}

\begin{CCSXML}
<ccs2012>
   <concept>
       <concept_id>10010147.10010178.10010224.10010225.10010228</concept_id>
       <concept_desc>Computing methodologies~Activity recognition and understanding</concept_desc>
       <concept_significance>500</concept_significance>
       </concept>
   <concept>
       <concept_id>10010147.10010257.10010293.10010294</concept_id>
       <concept_desc>Computing methodologies~Neural networks</concept_desc>
       <concept_significance>500</concept_significance>
       </concept>
   <concept>
       <concept_id>10010147.10010257</concept_id>
       <concept_desc>Computing methodologies~Machine learning</concept_desc>
       <concept_significance>500</concept_significance>
       </concept>
   <concept>
       <concept_id>10010147.10010178.10010224.10010225.10010227</concept_id>
       <concept_desc>Computing methodologies~Scene understanding</concept_desc>
       <concept_significance>300</concept_significance>
       </concept>
       <concept>
    <concept_id>10002944.10011122.10002945</concept_id>
    <concept_desc>General and reference~Surveys and overviews</concept_desc>
    <concept_significance>500</concept_significance>
</concept>
 </ccs2012>
\end{CCSXML}

\ccsdesc[500]{General and reference~Surveys and overviews}
\ccsdesc[500]{Computing methodologies~Activity recognition and understanding}
\ccsdesc[300]{Computing methodologies~Scene understanding}
\ccsdesc[300]{Computing methodologies~Neural networks}

\keywords{Multimodal, action recognition, fusion, deep learning, neural networks}

% \received{13 March 2023}
%  \received[revised]{12 March 2023}
%  \received[accepted]{5 June 2023}

\maketitle

\input{1.intro}

\input{2.learning-methods}

\input{4.datasets}

\input{3.fusion-methods}
\input{5.challenge}
\input{6.conclusion}

\begin{acks}
This work is partially funded by Edith Cowan University (ECU) and the Higher Education Commission (HEC) of Pakistan under Project \#PM/HRDI-UESTPs/UETs-I/Phase-1/Batch-VI/2018. Naveed Akhtar is the recipient of an Australian Research Council Discovery Early Career Researcher Award (project number DE230101058) funded by the Australian Government. 
\end{acks}

\bibliographystyle{ACM-Reference-Format}
\bibliography{main}

% \section*{Appendix}

% \begin{table}[b]
% \caption{List of acronyms and corresponding definitions}
% \label{tab:acronyms}
% \begin{tabular}{ll}
% \toprule
% \textbf{Acronyms} & \textbf{Definitions}          \\ \midrule
% AcR & Action Recognition \\
% CA & Cross Attention\\ 
% CAF & Cross-Attention Fusion \\
% CoAF & Co-Attention Fusion \\
% CNNs & Convolutional Neural Networks\\ 
% EF & Encoder Fusion \\
% FC & Fully Connected \\
% FFN & Feed-Forward Network \\
% GPU & Graphics Processing Unit\\ 
% HEF & Hierarchichal Encoder Fusion \\
% IR & Infrared \\
% LRCN & Long-term Recurrent Convolutional Network\\ 
% LSTM & Long-Short Term Memory\\ 
% mAP & mean Average Precision\\ 
% MAT & Multimodal Attention \\
% MHA & Multi-Head Attention \\
% MHAR & Multimodal Human Action Recognition   \\ 
% NN & Neural Network \\
% OOD & Out-of-Distribution \\
% PE & Positional Encodings \\
% PR & Pattern Recognition \\
% RGB & Red Green Blue  \\
% RGB+D & Red Green Blue+Depth \\
% ReLU & Rectified Linear Unit \\
% RNNs & Recurrent Neural Networks\\ 
% SA & Self-Attention \\
% SVM & Scalar Vector Machine \\
% Toyota-SH & Toyota-Smarthome \\
% VATN & Video Action Transformer Network \\
% VTs & Vision Transformers \\

% \bottomrule
% \end{tabular}
% \end{table}

\end{document}

%% file: 0.abstract.tex
\begin{abstract}
Due to its widespread applications, human action recognition is one of the most widely studied research problems in Computer Vision. Recent studies have shown that addressing it using multimodal data  leads to superior performance as compared to relying on a single data modality. During the adoption of deep learning for visual modelling in the last decade, action recognition approaches have mainly relied on Convolutional Neural Networks (CNNs). However, the recent rise of Transformers in visual modelling is now also causing a paradigm shift for the action recognition task. This survey captures this transition while focusing on Multimodal Human Action Recognition (MHAR). Unique to the induction of multimodal computational models is the process of `fusing' the features of the individual data modalities. Hence, we specifically focus on the fusion design aspects of the MHAR approaches. We analyze the classic and emerging techniques in this regard, while also highlighting the popular trends in the adaption of CNN and Transformer building blocks for the overall problem. In particular, we emphasize on recent design choices that have led to more efficient MHAR models. Unlike existing reviews, which discuss Human Action Recognition from a broad perspective, this survey is specifically aimed at pushing the boundaries of MHAR research by identifying promising architectural and fusion design choices to train practicable models. We also provide an outlook of the multimodal datasets from their scale and evaluation viewpoint. Finally, building on the reviewed literature, we  discuss the challenges and future avenues for MHAR.
\end{abstract}

%% file: 1.intro.tex
\section{Introduction}
\label{one}

\begin{figure*}
    \centering    
    \includegraphics[width=1\linewidth]{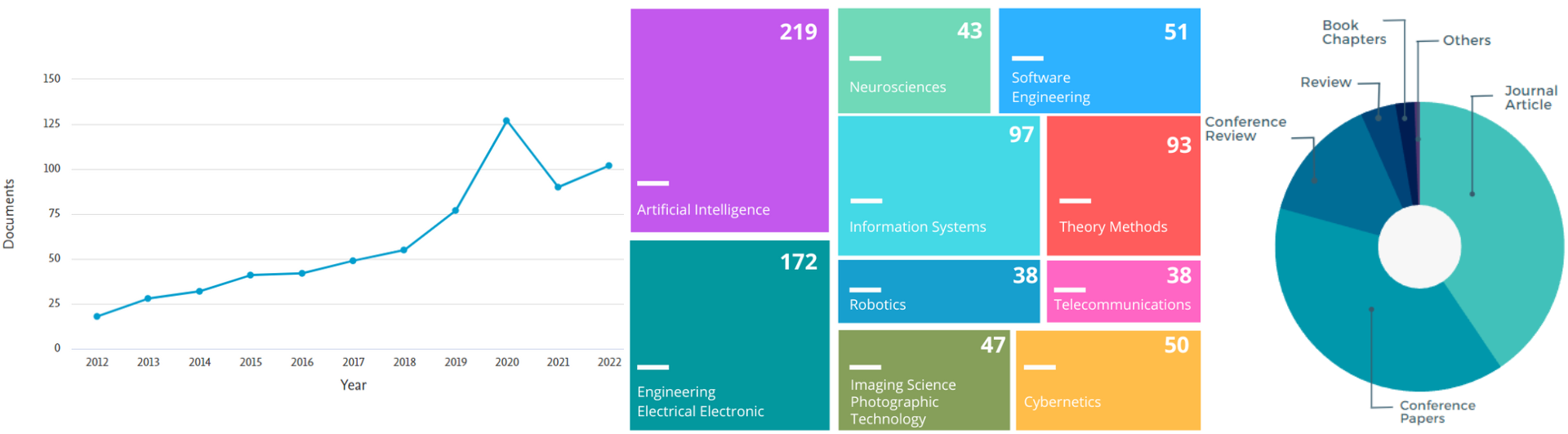}
    %combined-canvas-sc-b.png}
    \caption{(\textbf{Left}) Number of relevant publications in recent years, identified with the data collected from the Web of Science. (\textbf{Center}) Categories of publication contributions to different sub-fields of Science - generated with data from the Web of Science. (\textbf{Right}) Distribution as document type (data collected from Scopus). %Category `Others' includes Conference Review, Short Survey, Book, Note, Editorial, Letter, Erratum, Business Article, Data Paper, Retracted, Abstract Report, Report, and Undefined. %All categories which are less than 1\% individually have been grouped together.
    }
    \label{fig:combinedchart}
\end{figure*}

Modality broadly refers to the mode in which information is perceived. Humans are able to see, feel, hear and so on, using their senses. Our ability to intelligently process this multimodal information is what makes us highly effective beings. Currently, the field of  Artificial Intelligence (AI) is generally concerned with developing capabilities for individual data modalities that encode specific information about our surroundings. In this regard,  in the form of recent breakthroughs such as  GPT-4 \cite{openai2023gpt4} for multimodal 
 data (image and text), ChatGPT for text, DALL-E 2~\cite{ramesh2021zero} and Stable Diffusion \cite{rombach2022high} for the visual data, we are already witnessing human-level intelligence of machines. However, the ultimate objective to achieve Artificial General Intelligence (AGI) demands much more. The ability to handle and leverage information encoded in multimodal data is one of the basic needs of AGI. With the availability of increasingly powerful computational resources, this realization is fast re-writing the guiding principles for the research communities in different sub-fields of AI.  

In the domain of Computer Vision, human action recognition is a long-standing problem~\cite{Wang_2019_CVPR_Workshops}. This article follows Herath et al.~\cite{HERATH20174}, treating human action as  “the most elementary human-surrounding interaction with a meaning.” Human action recognition is thus the automated labelling process of human actions within a given sequence of visual frames. Due to its widespread practical applications, ranging from safety~\cite{sun2022safety} to healthcare \cite{zlatintsi2020support}, particularly for COVID \cite{ToMM2021Rahman} and many other downstream tasks \cite{fassold2019towards, qi2019stagnet, feng2017learning}, human action recognition has historically received significant attention within the vision research community. Recent years have also seen the emergence of numerous multimodal  approaches for this task, which define the research direction of Multimodal Human Action Recognition (MHAR).  Multimodal data fusion refers
to the mixing of features from data of various modalities. Fusion analysis has been extensively used in several domains including activity recognition \cite{ToMM2017Min}, 3D shape classification \cite{ToMM2020Nie} and predicting human eye fixation \cite{ToMM2020Yi}, as evident through relevant works.

%\IEEEPARstart{T}{he} world surrounding us is comprised of multiple modalities--- we see visuals, feel touch, hear audios, and so on. Commonly, modality refers to the way how something is observed or experienced. Therefore, when something is observed is multiple ways, then it is considered as multimodal. The vision of integration of natural language processing systems such as ChatGPT \cite{zhang2023modeling} into various computer vision domains for example, DALL-E 2\cite{ramesh2021zero} and stable diffusion \cite{rombach2022high} are accelerating the usage of advanced deep learning methods along with multimodal data for achieving artificial general intelligence (AGI). \par

%For the artificial intelligence of the future, it is important to comprehend the world surrounding us and reason events using multimodal messages. Models that are based on features extracted from diverse modalities are more robust than features from a single modality. Action Recognition is an interesting research area in the field of computer vision. Multimodality brings a new perspective to action recognition problem. In multimodal datasets, each modality adds complementary information to learning framework, which makes the overall recognition more robust and diverse.  \par

Presently, MHAR systems have been explored in various real-world scenarios of indoor and outdoor settings that relate to the application domains of healthcare \cite{   bruce2020vision, zahin2019sensor}, smart homes \cite{sharif2020human,nan2019human, liciotti2020sequential}, surveillance systems \cite{ ullah2019action, khan2020human, batchuluun2019action}, social relations recognition \cite{ToMM2021Xu} and numerous  other domains \cite{susan2019edge, tingting2019age}. It is evident from the existing literature that owing to the diversity of data, the research direction of MHAR has its own unique challenges and opportunities when compared to conventional action recognition tasks  \cite{mohsen2022artificial}. To contextualize this, early research in human action recognition was limited to only analysing still color images or videos \cite{guo2014survey,zhu2016intro}. The MHAR problem demands accounting for different data modalities while providing the opportunity for improved performance due to the availability of complementary information in different modalities.  Nevertheless, it is worth noting here that the typical challenges of classical action recognition, (e.g., background clutter, partial object occlusion, viewpoint variations, lighting changes, and execution rate etc.) remain equally pertinent to MHAR techniques.  

This survey focuses on MHAR approaches that use computer vision and signal-processing techniques to recognize human actions. Generally, techniques based on multimodal data are motivated based on the expectation to achieve better performance than unimodal data techniques. However, MHAR research has also shown other motivational sources in the form of the development of cost-effective sensing (e.g., ASUS Xtion \cite{xtion}, Microsoft Kinect \cite{Kinect}, and Intel RealSense \cite{RealSense}) and widespread gain in the computational power. Combined, these factors currently propel MHAR research to an unprecedented pace.
Accordingly, there has been a growing number of MHAR-related publications in the last few years - see  Fig.~\ref{fig:combinedchart} (left). Interestingly, these publications have contributed to diverse sub-fields of Science as their application domains - Fig. \ref{fig:combinedchart} (middle). 
%Most studies have been conducted by groups of researchers with a range of expertise and experience and published in top-ranked journals and conferences. 
In Fig \ref{fig:combinedchart}, we also show the Scopus article-type distribution against the query ``multimodal action recognition" to provide an overview of the publication trends in this highly active research direction.%, divided across fifteen different fields, based on the count, which is the total number of articles published in the last decade in a particular field. The count also includes the articles that are fully peer-reviewed, citable, and published but have not assigned a volume/issue/page number. 
%In Fig. \ref{fig:combinedchart}, the distribution of publications is presented as per Scopus\footnote{https://www.scopus.com/} defined document types. \par

Due to the emerging popularity of the topic, several survey articles have examined MHAR from different perspectives~\cite{atrey2010multimodal, ToMM2021wang, ToMM2019Zhang, ToMM2019Ahmad,wang2018rgb,aggarwal2014survey,shaikh2021rgb, HAN2017skeletal,zhang2016rgb,fan2016survey, Zhang2019,chen2017survey,zhang2017survey, Sun_2022}. 
However, this article is distinct from all previous studies, as we specifically focus and extensively cover the area of multimodal fusion and learning in action recognition. %Several survey papers \cite{beddiar2020vision,wang2018rgb,aggarwal2014survey,shaikh2021rgb,HAN2017skeletal,zhang2016rgb,fan2016survey,Zhang2019,chen2017survey,zhang2017survey} have discussed MHAR from different perspectives.
To the best of our knowledge, no prior detailed review article has thoroughly addressed this research direction. 
The specific contributions of this survey are summarized as follows. 
\begin{itemize}
\item To the best of our knowledge, this is the first comprehensive review of MHAR methods from the perspective of multimodality, with a focus on CNNs and transformer-based techniques.
%\item We comprehensively review, multimodal HAR methods, and categorize them into CNN and transformer-based approaches.        
\item It provides a thorough review of fusion approaches beyond the typical early and late splits used in the implementation of multimodal action recognition.  
\item It provides a comprehensive comparison of the existing methods and their performance on benchmark datasets, along with insightful discussions.
\item It pays special attention to the more recent approaches for MHAR, thereby  providing readers with an accessible overview of the current state-of-the-art in MHAR.

\end{itemize} 

\begin{figure*}
    \centering
    \includegraphics[width=0.8\textwidth]{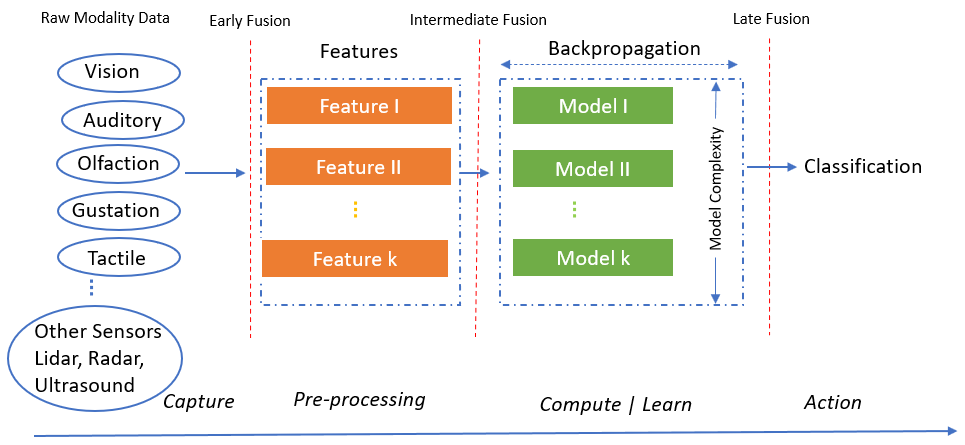}
    \caption{A typical multimodal fusion-based action recognition pipeline.}
    \label{fig:mhar-pipeline}
\end{figure*}

% The structure of this article is as follows. Section~\ref{two} introduces multimodal learning methods while discussing the original CNN and Transformer architectures and several methods derived using both of these architectures.  Standard benchmark datasets and corresponding state-of-the-art are summarized in Section~\ref{four}. Section~\ref{three} delves into fusion mechanisms in multimodal action recognition methods in depth.  Whereby, we review recent advances in different multimodal action learning architectures using various multimodal fusion strategies.  Section~\ref{five} discusses the overall challenges across the current multimodal action recognition spectrum. Conclusions are discussed in Section \ref{six}. 

The article is organized as follows: Section~\ref{two} discusses multimodal learning methods, covering the original CNN and Transformer architectures and their derivatives. Section~\ref{four} summarizes standard benchmark datasets and their state-of-the-art. Fusion mechanisms in multimodal action recognition are explored in Section~\ref{three}, highlighting recent advancements in various fusion strategies. Section~\ref{five} addresses the challenges in current multimodal action recognition. Section \ref{six} concludes the article.
% A list of acronyms used throughout this
% article is provided in Table \ref{tab:acronyms}.

%% file: 2.learning-methods.tex
\section{Multimodal Learning Methods}
\label{two}

\subsection{Multimodal Learning}

Before discussing multimodal learning, we must first understand unimodal learning, where a neural network processes a single data type.

Given a dataset \(T = \{{x_1,...,x_n, y_1,...,y_n}\}\), \(x_i\) is the i-th training example and \(y_i\) is the true label of \(x_i\). Training on a single modality, say RGB images, can be formalized by the equation:
\begin{equation}
L(C (\phi_m(X)), y)
\end{equation}
\noindent where \(\phi_m\) is a deep neural network tailored for that modality (like a CNN for images). Its parameters are represented by \(\ominus_m\). \(C\) is a classifier that predicts the label based on the features extracted by \(\phi_m\). This classifier typically uses one or more fully connected layers, and its parameters are denoted by \(\ominus_c\).

When dealing with real-world problems, often a single data modality is not enough. For example, in video understanding tasks, both video frames (RGB) and audio can provide valuable information. By leveraging multiple modalities, a model can potentially achieve better performance than relying on a single one. A typical multimodal fusion-based system would follow a standard workflow (see Fig. \ref{fig:mhar-pipeline}).

Training with multiple modalities can be formalized as:
\begin{equation}
L_{\textrm{multi}} = L(C ( \phi_{\textrm{audio}}\oplus\phi_{\textrm{video}} ), y)
\end{equation}

\noindent here, each modality, audio and video, has its respective deep neural network, \(\phi_{\textrm{audio}}\) and \(\phi_{\textrm{video}}\), designed to extract relevant features. The fusion operation, represented by \(\oplus\), can be a simple concatenation, or more sophisticated operations like weighted sums or attention mechanisms.

\subsection{Learning Methods in CNNs}

\subsubsection{Convolution Neural Networks (CNNs)}

CNN architectures found in the literature are diverse while sharing common elements. Primarily, they incorporate convolutional layers, where multiple kernels compute various feature maps based on input data. Generating a new feature map involves convolving the input with a learned kernel and applying an activation function to the results. Shared kernels, applied across all spatial input locations, produce individual feature maps, simplifying model complexity and improving trainability. The feature value at a specific location $(i, j)$ in the $k$-th feature map of the $l$-th layer, denoted as $z^l_{i, j, k}$, is given by the equation:
\begin{equation}
z^l_{i, j, k} = {w^l_k}^T x^l_{i, j} + b^l_k
\end{equation}
Here, the shared kernel $w^l_k$ generates the feature map $z^l_{:,:,k}$. The activation function $a(\cdot)$ introduces crucial non-linearities, determining the convolutional feature activation value $a^l_{i, j, k}$ as follows:
\begin{equation}
a^l_{i, j, k} = a(z^l_{i, j, k})
\end{equation}

The pooling layer enhances shift-invariance by reducing feature map resolution. For each feature map $a^l_{:,:,k}$, the pooling function, denoted as pool$(\cdot)$, is applied as follows:
\begin{equation}
y^l_{i, j, k} = \text{pool}(a^l_{m, n, k}), \forall (m, n) \in \mathbb{R}_{ij}   
\end{equation}

After multiple convolutional and pooling layers, fully connected layers may exist for high-level reasoning. The output layer, often employing the softmax operator for classification, serves as the final layer. CNN parameters, denoted by $\theta$, including weight vectors and bias terms, are optimized by minimizing a task-specific loss function. For $N$ input-output pairs ${(x^{(n)}, y^{(n)} ); n \in [1, N]}$, the CNN loss is calculated as:
\begin{equation}
\mathcal{L}_{\textrm{multi}} = \frac{1}{N} \sum_{n=1}^{N} \lambda(\theta ; y^{(n)}, o^{(n)} )
\end{equation}

\subsubsection{CNN-based Learning Methods}

\begin{table*}[]
\caption{Summary of CNN-based MHAR methods applied on standard benchmark public datasets using fusion methods. Abbreviations: Modality (S: Skeleton, IR: Infrared, P: Pose,  D: Depth), Datasets (U: UTD-MHAD, UT: UT-Kinect, N1:NTU RGB+D 60, N2:NTU RGB+D 120, NW: NW-UCLA, T: Toyota SH, M: MSAR Daily, UCF: UCF51).}
\label{tab:datatech}
\centering
\begin{tabular}{@{}lcccccccccc@{}}
\toprule
\textbf{Ref.} & \textbf{Modality} & \textbf{Fusion} & \textbf{U} & \textbf{UT} & \textbf{N1} & \textbf{N2} & \textbf{NW} & \textbf{T} & \textbf{M} & \textbf{UCF} \\
\midrule
\cite{islam2020hamlet} & RGB+S & Middle & 95.12 & 97.56 & - & - & - & - & - & - \\
\cite{memmesheimer2020gimme} & S+IR & Early & 93.3 & - & 70.8 & 78.3 & - & - & - & - \\
\cite{das2020vpn} & RGB+P & Middle & - & - & 95.5 & 93.5 & 86.3 & 60.8 & - & - \\
\cite{deboissiere2020infrared} & P+IR & Late & - & - & 91.6 & - & - & - & - & - \\

\cite{joze2020mmtm}     & RGB+P & Middle & - & - & 91.99 & - & - & - & - & - \\

\cite{Perez-Rua_2019_CVPR} & RGB+P & Hybrid & - & - & 90.04 & - & - & - & - & - \\

\cite{Das_2019_ICCV}  & RGB+P & Late & - & - & 92.2 & - & 90.1 & 54.2 & - & - \\

\cite{Liu_2018_CVPR} & Heatmap+P & Late & 94.5 & - & 91.7 & - & - & - & - & - \\

\cite{zhu2018action}  & RGB+P & Late & 92.5 & - & 94.3 & - & - & - & - & - \\

\cite{fabian2018clouds}  & S+P & Late & - & - & 86.6 & - & - & - & - & - \\

\cite{luvizon2018pose} & RGB+P & Late & - & - & 85.5 & - & - & - & - & - \\

\cite{KHAIRE2018107} & RGB+S+D & Late & 95.1 & - & - & - & - & - & - & - \\

\cite{DBLP:journals/corr/ShahroudyNGW16} & RGB+D & Middle & - & - & 74.9 & - & - & - & 97.5 & - \\

\cite{shaikh2022maivar,shaikh2023ncaa} & RGB+Audio & Middle & - & - & - & - & - & - & - & 86.7 \\
\bottomrule
\end{tabular}
\end{table*}

Convolutional Neural Network (CNN)-based Multimodal Human Activity Recognition (MHAR) approaches can leverage the potential of automated feature learning, wherein varying modalities inform and interact with each other through concealed features. For a comparison of these techniques on benchmark datasets, please refer to Table \ref{tab:datatech}. Some significant studies are summarized below. \par 

%\paragraph{Multi-stream approaches}

Islam et al. \cite{islam2020hamlet} introduced a Hierarchical Multimodal Attention-based Human Activity Recognition Algorithm (HAMLET), which uses a unique feature encoder and a multi-head self-attention mechanism for each modality to encode spatio-temporal features. HAMLET employs a multimodal self-attention-based fusion architecture, called Multimodal Atention-based Feature
Fusion (MAT), which blends attention-based fused features with the use of sum (extracted unimodal features are summed
after applying multimodal attention) and concatenate (in this approach the attended multimodal features are concatenated) operations. Action classification is done using a fully connected layer that harnesses the computed multimodal features. Contrastingly, \cite{memmesheimer2020gimme} developed an intuitive methodology that blends different modalities using a matrix concatenation operation, transforming signals into an image for classification via a 2D CNN. \par

Another distinctive approach put forth by \cite{das2020vpn}, combines spatial embeddings of RGB images and 3D poses using an attention mechanism for extracting superior discriminatory spatio-temporal patterns. Moreover, \cite{deboissiere2020infrared} has proposed the late fusion of feature vectors from infrared (IR) and 3D pose modules, classified by a multi-layer perceptron. Notably, \cite{joze2020mmtm} introduced an innovative method using intermediate fusion among RGB and 3D pose information through a Multimodal Transfer Module (MMTM). Moreover, \cite{luvizon2018pose} suggests a multi-task framework for concurrent 2D and 3D estimation from still images and HAR from video sequences. \par

The work proposed by \cite{Perez-Rua_2019_CVPR} used a network architecture search-based approach that applies a progressive algorithm for multi-fusion architecture search and the introduction of new fusion layers. This technique executes an explicit fusion of video and pose modalities via a search algorithm. Alternatively, \cite{Das_2019_ICCV} introduced an approach that utilizes a separable spatio-temporal attention model and a pose-driven attention model, late fusing scores by averaging softmax scores. \par

The Evolution of Pose Estimation Maps (PEM), proposed by \cite{Liu_2018_CVPR}, employs spatial rank pooling to aggregate the evolution of heatmaps as a body shape evolution image, and body-guided sampling for aggregating the evolution of poses as a body pose evolution image. Complementary features of both images are then probed through CNNs for action classification. Further, \cite{fabian2018clouds} has proposed a method that uses unstructured collections of spatio-temporal glimpses with distributed recurrent tracking. Additionally, \cite{DBLP:journals/corr/ShahroudyNGW16} has presented a deep multimodal feature analysis-based learning machine that applies mixed norms for component regularization and group selection for superior classification performance. Lastly, \cite{KHAIRE2018107} has used a 5-CNN-streams approach, based on Motion History Image (MHI), Front Depth Motion Maps (DMMs), Side DMM, Top DMM, and Skeleton images, fused at the decision level for action classification. \par
These various approaches offer different methods of fusing multimodal information and capitalizing on their synergies for improved performance in Human Activity Recognition. They demonstrate versatility and innovation in this rapidly advancing field.

\subsection{Learning Methods in Transformers}

We first introduce Transformers here and then illustrate the major trends seen across all Transformer-based architectures that have also been adopted for action recognition. 

\subsubsection{Transformers}

\begin{figure}
    \centering
    \includegraphics[height=0.5\textwidth,width=0.6\textwidth]{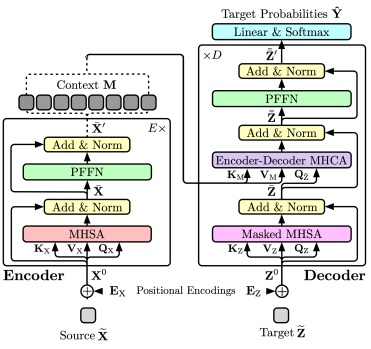}
    \caption{The Transformer, as originally proposed in \cite{vaswani2017attention}, depicted through visualization \cite{selvasurvey2022videotrans}.}
    \label{fig:transformer}
\end{figure}

Transformers, as introduced by Vaswani et al.~\cite{vaswani2017attention}, have redefined neural network architectures, primarily due to their superior capability in capturing long-term sequential data dependencies. The Transformer  was first proposed as a remedy to some limitations of sequence modeling architectures, originally designed to deal with whole sequences at once (see Fig. \ref{fig:transformer}), allowing parallelization of some operations (as opposed to RNNs, which are sequential in nature), and reducing the locality bias of traditional networks (such as CNNs). The key innovation in transformers is the self-attention mechanism, which allows the network to selectively focus on different parts of the input sequence based on their relevance to the current task. Different from RNNs, which operate sequentially, and traditional networks like CNNs that have a locality bias, Transformers treat entire sequences in parallel, a major breakthrough enabled by the self-attention mechanism.

The essence of the self-attention mechanism is computing three vectors for every input element: query, key, and value. The mechanism assigns weights to each input element based on its query vector's similarity, then derives an output from a weighted sum of the value vectors. Additionally, Transformers have other modules such as: \textbf{Multi-head attention}: It focuses on multiple sequence parts simultaneously by generating multiple sets of query, key, and value vectors. The outputs are then amalgamated using a linear operation.
    \begin{align}
    \text{Attention}(Q,K,V) &= \text{softmax}\left(\frac{QK^T}{\sqrt{d_k}}\right)V \\
    \text{MultiHead}(Q,K,V) &= \text{Concat}(head_1,\ldots,head_h)W^O
    \end{align}
    where 
    \begin{equation}
    head_i = \text{Attention}(QW_i^Q,KW_i^K,VW_i^V)
    \end{equation}
    and $W_i^Q$, $W_i^K$, $W_i^V$, and $W^O$ are the learnable weight matrices. \textbf{Position-wise feedforward network}: This module applies individual feedforward networks to every sequence element, which enhances the model's sequential data handling.
    \begin{equation}
    \text{Position-wise FeedForward}(x) = \text{max}(0,xW_1+b_1)W_2+b_2
    \end{equation}
    where $x$, $W_1$, $W_2$, $b_1$, and $b_2$ are input and learnable parameters.

\subsubsection{Transformers-based Learning Methods}

Transformer-based methods have also been explored for multimodal action recognition and have shown promising results. The self-attention mechanism of the Transformer allows it to selectively attend to different modalities and capture their temporal and spatial relationships, leading to improved performance compared to traditional methods \cite{gabeur2020multi}. Additionally, the use of cross-modal attention can help the model fuse information from different modalities and improve its overall accuracy \cite{Zhuang_2017_CVPR}.
 Table. \ref{tab:tlearn-methods} presents summary on architectures of some significant transformer-based works in MHAR and refers to various features like Normalization (Norm.), Activation (Act.), Encoder configuration (Heads/Layers), Efficiency (Aggregation, Restriction, and Weight-sharing), Positional-Encoding (PE, with type and strategy), and Backbone networks. 
 
Initial methods such as VATNet \cite{girdhar2019video}, CBT \cite{wang2019efficient}, and ELR \cite{purwanto2019extreme} set a precedent. Their successes paved the way for a surge in transformer-based approaches, revealing a shift from traditional CNN methods, which primarily hinged on spatial features. Most methods either use Pre or Post normalization. When paired with activation, GeLU and ReLU seem to be dominant choices. The varied combinations indicate exploration to pinpoint the most efficient configuration.

The hierarchy of encoder architecture varies across different methods. While some like Actor-T \cite{gavrilyuk2020actor} showcase simpler design configurations, others like Video Swin \cite{liu2022video} take a deep dive with relatively complex design choices. This reflects the diverse depth of transformer models tailored for action recognition. However, notable methods, such as TimeSformer \cite{bertasius2021space}, ViViT \cite{arnab2021vivit}, VATT \cite{akbari2021vatt}, MViT \cite{fan2021multiscale}, SCT \cite{zha2021shifted}, and others, have indicated the inclusion of aggregation, restriction, and weight-sharing strategies for achieving computational efficiency in design.   

A majority of the methods, including ViViT \cite{arnab2021vivit}, SCT \cite{zha2021shifted}, and VTN \cite{wu2021rethinking}, adopt positional encoding mechanisms, with strategies ranging from Learnt (L) to Fixed (F) and from Absolute (A) to Relative (R). This underscores the essential role of temporal awareness in videos. For backbone choices, transformer-based methods for MHAR leverage diverse CNN-based backbones to capture spatiotemporal features essential for video understanding. In particular, I3D \cite{carreira2017quo} is commonly adopted, known for its efficacy in video-related tasks, as seen in models like VATNet \cite{girdhar2019video}, LapFormer \cite{kondo2021lapformer}, and Actor-T \cite{gavrilyuk2020actor}. S3D \cite{xie2017rethinking} in CBT \cite{wang2019efficient} efficiently separates spatial and temporal information, enhancing video content understanding. ResNet-50, utilized in LapFormer \cite{kondo2021lapformer}, LTT \cite{kalfaoglu2020late}, PEMM \cite{lee2020parameter}, TRX \cite{perrett2021temporal}, and GroupFormer \cite{li2021groupformer}, provides a strong foundation with residual connections, addressing the vanishing gradient problem. R(2+1)D \cite{tran2018closer}, found in LTT \cite{kalfaoglu2020late}, STiCA \cite{patrick2021space}, and GroupFormer \cite{li2021groupformer}, decomposes 3D convolutions for efficient spatiotemporal processing. SlowFast \cite{feichtenhofer2019slowfast}, employed in PEMM \cite{lee2020parameter} and CATE \cite{sun2021composable}, captures both slow and fast motion features, suitable for varied temporal dynamics. The use of HRNet \cite{wang2020deep} in Actor-T \cite{gavrilyuk2020actor} emphasizes maintaining high-resolution representations for detailed visual information. Additionally, various methods, including TimeSformer \cite{bertasius2021space}, ViViT \cite{arnab2021vivit}, FAST \cite{yu2021frequency}, VATT \cite{akbari2021vatt}, MViT \cite{fan2021multiscale}, SCT \cite{zha2021shifted}, CATE \cite{sun2021composable}, Video Swin \cite{liu2022video}, and VTN \cite{wu2021rethinking}, employ custom linear layers, showcasing a flexible approach to multimodal feature fusion within the Transformer architecture. Overall, the selection of CNN-based backbones reflects a strategic choice to enhance the ability of the model to recognize human actions in diverse multimodal contexts.

Transformers, despite their successes, struggle with challenges. A scarcity of large-scale multimodal datasets reduces their performance, making valid comparisons with CNN-based methods difficult. Their intricate architecture necessitates high computational power and memory. This underlines an acute need for streamlined transformer architectures.
The success of transformer-based models, driven by their flexibility and depth, has steered the action recognition domain into an experimental phase. Their departures and incorporations from CNN techniques not only notice their distinctive design philosophy but also underline the evolutionary trajectory of video understanding. However, addressing their inherent challenges remains pivotal for yielding their full potential in real-world scenarios.

 % \begin{landscape}
\begin{table}
\caption{Summary of significant Transformer-based architectures in MHAR. Abbreviations: Norm: Normalization, Act: Activation (Pre or Post), Encoder: number of Heads/Layers (``-'' subsequent blocks and ``*'' repeated modules), Efficient: ``X'' for indicating Aggregation,
Restriction and Weight-sharing, PE: Positional-Encoding (L: Learnt/F: Fixed, A:Absolute/R: Relative), T: Tokenization (Patch, Clip, and Frame levels), Backbone (ResNet and
DenseNet is abbreviated). Modality: (Visual, Audio, or Textual), SS: Self-Supervision,
NA: not available, X/``–'' applies or not.}
\label{tab:tlearn-methods}
\begin{tabular}{crlllllllllll}
\toprule
\textbf{Model} &  
  \textbf{Norm.} &
  \textbf{Act.} &
  \textbf{\begin{tabular}{@{}l@{}}Encoder \\(H/L)\end{tabular}} &
  %\textbf{Efficient } &
  \textbf{Backbone} & \textbf{PE} &  \textbf{T} &  \textbf{Modal.} &  \textbf{SS}  \\ \midrule
  
 VATNet\cite{girdhar2019video} &  Post  & ReLU & 2/3  & \begin{tabular}{@{}l@{}}I3D \cite{carreira2017quo} \\ Faster R-CNN  \cite{ren2015faster}\end{tabular} & FA & P & X – – & – \\
 CBT \cite{wang2019efficient} &  Post  & GeLU & 4-1/1-NA   &S3D \cite{xie2017rethinking} & – & C &  X – X & X \\
 ELR \cite{purwanto2019extreme} &  Post  & ReLU & 8/1   & I3D \cite{carreira2017quo} & – & P & X – – & – \\
 LapFormer \cite{kondo2021lapformer} &  Post  & ReLU & 4/3  & RN-50 \cite{he2016deep} & FA & P &  X – – & – \\
 LTT \cite{kalfaoglu2020late} & Pre  & GeLU & 8/1  &  R(2+1)D \cite{tran2018closer} & LA &  F & X – – & – \\
  Actor-T \cite{gavrilyuk2020actor} &  Post  & ReLU & 1/1  &  \begin{tabular}{@{}l@{}}I3D \cite{carreira2017quo} \\ HRNet \cite{wang2020deep} \end{tabular}  & FA & P & X – – & – \\
 
 TimeSformer  \cite{bertasius2021space}   &  Pre                 & GeLU & 12/12       & Linear Layer  & LA & P & X - - & -  \\ 
 PEMM \cite{lee2020parameter} &   Post  & GeLU & 6/12*3  &  \begin{tabular}{@{}l@{}}SlowFast \cite{feichtenhofer2019slowfast} \\ RN-50 \cite{he2016deep}\end{tabular} & LA & C & X X – & X \\ 
 ViViT \cite{arnab2021vivit} &  Pre  & GeLU & 16/24  & Linear Layer & LA & P & X – – & – \\
 
 FAST \cite{yu2021frequency} &  Pre  & GeLU & NA/8   & 2 Linear Layers & LA & P &  X – – &  – \\
 
 VATT \cite{akbari2021vatt} &   Pre  & GeLU & 12/12   & Linear Layer & LA & P & X X X & X \\
 MViT \cite{fan2021multiscale} &  Pre  & GeLU & \begin{tabular}{@{}l@{}}1-2-4-8/ \\ 1-2-11-2 \end{tabular} & Linear Layer & LA & P & X  – – & – \\
 SCT \cite{zha2021shifted} &  Pre  & GeLU & \begin{tabular}{@{}l@{}}4-8-8-8/ \\6-1-1-4\end{tabular}   & Linear Layer & LA & P & X – – & – \\
 CATE \cite{sun2021composable} & NA &  NA & 12/4   & SlowFast \cite{feichtenhofer2019slowfast} & – & C & X – – & X \\
 TRX \cite{perrett2021temporal}  & Pre  & NA & 1/1   & RN-50 \cite{he2016deep} & FA & F & X – – & – \\
  STiCA \cite{patrick2021space}  & Post  & ReLU & 4/2  & \begin{tabular}{@{}l@{}}R(2+1)D-18 \cite{tran2018closer} \\ RN-9 \cite{he2016deep} \end{tabular} & LA  & C & X X – & X \\
 GroupFormer \cite{li2021groupformer} &   Post  & ReLU & 8*4  & I3D \cite{carreira2017quo} & LA & F & X – – & – \\
 Video Swin  \cite{liu2022video} & Pre  & GeLU & \begin{tabular}{@{}l@{}}3-6-12-24/ \\2-2-6-2\end{tabular}   & Linear Layer & LR & P & X – – & – \\
 MAiVAR-T \cite{shaikh2023maivart} & Pre  & ReLU & \begin{tabular}{@{}l@{}}3-6-12-24/ \\2-2-6-2\end{tabular}   & TSN \cite{wang2016temporal} & - & P & X X – & – \\
 VTN \cite{wu2021rethinking}  & Pre  & GeLU & 12-12/12-3 & Linear Layer & LA & P & X – – & – \\
\bottomrule
\end{tabular}
\end{table}
  %\end{landscape}

%% file: 4.datasets.tex
\section{Datasets}
\label{four}

% A2

A multitude of datasets have been created to develop and evaluate multimodal human activity recognition (MHAR) methods. In Table \ref{tab:mmdatasets}, a series of benchmark datasets are listed along with their attributes such as scale, number of categories and provided modalities. For RGB-based MHAR, UCF101 \cite{soomro2012ucf101}, HMDB51 \cite{hmdb2011data}, and Kinectis-400 \cite{kay2017kinetics} are widely used as benchmark datasets, along with Kinetics-600 \cite{carreira2018short}, Kinetics-700 \cite{carreira2019short}, EPIC-KITCHENS-55 \cite{Damen2018EPICKITCHENS}, and ActivityNet \cite{caba2015activitynet}. Additionally, the datasets MSR-DailyActivity3D \cite{multiview} and Northwestern-UCLA \cite{wang2014cross} are widely used for depth-based MHAR.  Moreover, for MHAR, large benchmark datasets suitable for fusion and co-learning of different modalities include NTU RGB+D \cite{shahroudy2016ntu}, NTU RGB+D 120 \cite{Liu_2019}, and MMAct \cite{kong2019mmact}. The datasets UTD-MHAD \cite{utd2015data}, and PKU-MMD \cite{liu2017pku} are also popularly used. \par

 MHAR datasets are generally created using different sensor perspectives and settings. For instance, there are single-view action datasets, multi-view action datasets and multi-person/interaction action datasets. In the single-view action datasets, a specific single viewpoint is captured. In the multi-view datasets, two or more viewpoints of a single action are captured in the frame sequences. In multi-person/interaction action datasets, action/activity is normally performed between two or more people \cite{zhang2016rgb}. Additionally, action recognition datasets are normally available in different modalities, namely RGB, depth videos, skeleton joint positions, inertial sensor signals, and IR. These  modalities often allow capturing different aspects of information  for the same action scenario. As can be seen in Table \ref{tab:mmdatasets}, the datasets also vary considerably in terms of the size and number of classes. As there is no public platform similar to YouTube available for MHAR dataset generation, MHAR currently lacks large-scale benchmark datasets when compared to video-based action recognition. \par

\begin{table*}[]
\caption{Public multimodal datasets. Notations : Seg: Segmented, Con: Continuous, D: Depth, S: Skeleton, Au: Audio, Ac: Accelerometer, IR: Infrared, \#: Number of. \textit{* action classes with $>$ 50 samples}\textcolor{red}. Note: Some RGB datasets are considered as multimodal because different modalities are from it such as audio, skeleton, and optical flow as well as some representations from compressed video formats.}
\label{tab:mmdatasets}
\resizebox{\textwidth}{!}{%
\begin{tabular}{lrlllllll}
\toprule
\textbf{Dataset} &
  \textbf{Year} &
  \textbf{Data Source} &
  \textbf{Type} &
  \textbf{Modality} &
  \textbf{\#Actions} &
  \textbf{\#Subjects} &
  \textbf{\#Samples} &
  \textbf{\#Views} \\ \midrule
CMU MoCap \cite{mocap}                        & \textquotesingle01 & MoCap                & Seg & RGB,S         & 45  & 144 & 2,235  & 1  \\ \hline
MSR-Action3D \cite{multiview}                 & \textquotesingle10 & Kinect v1            & Seg & S,D           & 20  & 10  & 567   & 1  \\ \hline
RGBD-HuDaAct \cite{ni2011rgbd}                                                   & \textquotesingle11 & Kinect v1            & Seg & RGB+D         & 13  & 30  & 1,189  & 1  \\ \hline
HMDB-51   \cite{hmdb2011data}  & \textquotesingle11     & YouTube                      &   Seg  &         RGB      &  51  &  -   &  6,766     &  1  \\ \hline
MSR-DailyActivity3D \cite{multiview}          & \textquotesingle12 & Kinect v1            & Seg & RGB,D,S       & 16  & 10  & 320   & 1  \\ \hline
UCF-101    \cite{soomro2012ucf101}  & \textquotesingle12     &    YouTube                  &  Seg   &     RGB     &  101  & -    &   13,320    &  1  \\ \hline
UT-Kinect \cite{xia2012view}          & \textquotesingle12 & Kinect v1            & Seg & RGB,D,S       & 10  & 10  & 200   & 1  \\ \hline
SBU Kinect \cite{sbukinect2012}  & \textquotesingle12 & Kinect v1            & Seg & RGB,D,S       & 7   & 8   & 300   & 1  \\ \hline
3D Action Pairs \cite{Oreifej2013depth}                                                & \textquotesingle13 & Kinect v1            & Seg & RGB,D,S       & 12  & 10  & 360   & 1  \\ \hline
Berkeley MHAD \cite{mhad}                     & \textquotesingle13 & MoCap + Kinect v1      & Seg & RGB,D,S,Au,Ac & 12  & 12  & 660   & 4  \\ \hline
NW-UCLA \cite{wang2014cross} & \textquotesingle14 & Kinect v1            & Seg & RGB,D,S       & 10  & 10  & 1,475  & 3  \\ \hline
UTD-MHAD \cite{utd2015data}                                                   & \textquotesingle15 & Kinect v1 + Inertial & Seg & RGB,D,S       & 27  & 8   & 861   & 1  \\ \hline
NTU RGB+D 60  \cite{shahroudy2016ntu}         & \textquotesingle16 & Kinect v2            & Seg & RGB,D,S,IR    & 60  & 40  & 56,880 & 80 \\ \hline
PKU-MMD \cite{liu2017pku}                     & \textquotesingle17 & Kinect v1            & Con & RGB,D,S,IR    & 51  & 66  & 1,076  & 3  \\ \hline

ActivityNet-200 \cite{caba2015activitynet}      & \textquotesingle16     &    YouTube                  &  Con  &      RGB,X          &   200  &   -  &   19,994    & 1  \\ \hline

Kinetics 400 \cite{kay2017kinetics}     &  \textquotesingle17    &  YouTube    &  Seg   &  RGB    &  400  &   -  &  306,245     & 1   \\ \hline

Kinetics 600 \cite{carreira2018short}    &  \textquotesingle18    &   YouTube   &  Seg   &  RGB    & 600   &   -  &  495,547   & 1   \\ \hline

AVE     \cite{tian2018audio}                           &   \textquotesingle18   &     YouTube    & Seg    &   RGB,Audio      &  28   & -  & 4,143     &  1  \\ \hline

EPIC Kitchen 55   \cite{Damen2018EPICKITCHENS}                          &  \textquotesingle18    &     GoPro Hero 5 &  Con   &  RGB    & 149*   & 32    & 39,596      &   1  \\ \hline

NTU RGB+D 120 \cite{Liu_2019} &
  \textquotesingle19 &
  Kinect v2 &
  Seg &
  RGB,D,S,IR &
  120 &
  106 &
  114,480 &
  155 \\ \hline

Toyota-SH    \cite{Das_2019_ICCV}                                           & \textquotesingle19 & Kinect v1            & Seg & RGB,D,S       & 31  & 18  & 
16,115 & 1  \\ \hline

Kinetics 700  \cite{carreira2019short}   &  \textquotesingle19    &   YouTube   &  Seg   &   RGB   & 700   &   -  & 650,317      &   1\\ \hline

EPIC Kitchen 100     \cite{damen2022rescaling}                           &   \textquotesingle20   &     GoPro Hero 7    & Con    &   RGB,Flow      &  4,053   & 37  & 89,977     &  1  \\ 
\bottomrule
\end{tabular}
}
\end{table*}

MHAR datasets contain continuous (Con) or segmented (Seg) videos as reported in Table \ref{tab:mmdatasets}. The continuous video  datasets comprise more than one action in a single video or frame sequence. They are mainly used for localization, detection, and prediction of actions. The techniques covered in this article generally focus on segmented datasets, which have complete actions in individual data instances. These datasets can be easily used for the evaluation of action recognition algorithms. Below, we discuss different aspects of commonly used segmented datasets. \par
The CMU MoCap \cite{mocap} is a graphics motion-capture database. It is one of the earliest sources of action data, that covers a variety of actions, including the interaction between two subjects, locomotion with uneven terrains, sports and several other human actions. CMU MoCap is a segmented dataset with 45 classes of action performed on 144 subjects containing 2,235 video samples. This dataset has a single line of sight, and it contains RGB and skeleton modalities. 
The MSR-Action3D \cite{multiview} is the first RGB+D action dataset captured by the Kinect sensors. It contains 20 actions performed by ten subjects. This dataset requires  some post-processing to remove the background. In particular, the right arms and legs are known to be preferred in this dataset for performing several related actions. 

The MSR-DailyActivity3D dataset \cite{multiview} is collected by Microsoft and Northwestern University,  focusing on daily activities in a living room. A total of 16 actions are  performed by ten subjects while sitting on a sofa or standing close to a sofa. This dataset functions as a good baseline  for multimodal video analysis, providing a small number of classes. The dataset provides RGB, depth and skeleton data modalities. \par
The Berkeley Multimodal Human Action Database (Berkeley MHAD) \cite{mhad} is the first dataset to be captured with five different modalities. One optical MoCap system, four multi-view cameras, two Kinect v1 cameras, six wireless accelerometers, and four microphones are used in the creation of this dataset. Berkeley MHAD contains eleven actions performed by seven male and five female subjects, each varying in terms of style and speed. These actions are  divided into the categories of: (1) full body movement actions, e.g., jumping jacks, throwing; (2) high dynamics in the upper extremity actions, e.g., waving hands, clapping hands; and (3) high dynamics in the lower extremity actions, e.g., sitting down, standing up. Audio and accelerometer data are two rare  modalities  in this dataset. 

The NTU RGB+D 120 \cite{Liu_2019} dataset, as highlighted in Table~\ref{tab:mmdatasets}, is currently one of the largest action recognition datasets in terms of the number of samples per action. It has RGB, depth, skeleton, and IR modalities, captured with Kinect sensor v2.  The dataset has more than 114,000 action video samples and up to 8 million image frames. The dataset contains 120 classes of daily and health-related actions performed by 106 subjects aged around 10 to 57 years, with different cultural backgrounds. Further, it consists of 155 different camera viewpoints. This dataset is a strong candidate for evaluating the scalability and multimodal nature of algorithms.

The Toyota-Smarthome (Toyota-SH) \cite{Das_2019_ICCV} dataset contains approximately 16,100 unscripted video samples with 31 action classes performed by 16 subjects. This dataset is one of the most recent action recognition datasets, with three different scenes and seven camera viewpoints. Toyota-SH dataset provides several real-world challenges such as improvisational acting, camera framing, composite activities, multi-view and the same activity using different objects. A unique feature of this dataset is that its actions are performed by subjects who did not receive any information regarding how to perform them. This dataset provides RGB, depth and skeleton modalities.

%% file: 3.fusion-methods.tex
\section{Fusion Methods}
\label{three}
To this end, recent
works have started to explore different fusing methods to incorporate multi-sensory data streams effectively. Compared with
typical CNNs, a Transformer is naturally appropriate for
multi-stream data fusion because of its nonspecific embedding
and dynamically interactive attention mechanisms. Multimodal data fusion refers to the mixing of features from data of various modalities. The objective of multimodal data fusion is to achieve better accuracy than a single modality. Data fusion supports diversity, enhancing the uses, advantages and analysis of ways that cannot be achieved through a single modality. Merging different modalities, as in \cite{ToMM2022yin}, have many benefits such as enhanced signal-to-noise ratio, improved confidence, increased robustness and reliability, enhanced resolution, better precision and discrimination as well as robustness against interference  \cite{antonio2019mmfusion}. 

In the context of MHAR, the choice of formulation depends on data characteristics and specific task requirements, necessitating experimentation and empirical evaluation to identify the most effective fusion strategy for a given multimodal dataset. Researchers explore diverse architectures and fusion methods to optimize accuracy and robustness in MHAR systems. Fusion techniques can be deployed at different stages in the action recognition process to acquire combinations of distinct information. These fusion approaches depend on the type of data and technique used for action recognition. Multimodal data fusion approaches could be classified into classical-machine-learning-based and deep learning-based, where latter can be further divided into CNNs- and Transformers-based, according to architectural choices. Various fusion approaches are discussed in the following subsections.

\subsection{Fusion in CNNs}

Data fusion in deep-learning-based MHAR techniques can be applied at similar stages as in classical-machine-learning-based approaches. In the context of fusion in MHAR, a common strategy involves employing modality-specific networks, where separate CNNs process each modality independently, and their outputs are combined through fully connected layers or other fusion techniques. Additionally, for video-based action recognition, 3D CNNs capture spatial and temporal features simultaneously, offering a natural approach to fuse information across space and time for multiple modalities.

For video-based multimodal action recognition, the utilization of 3D CNNs, which capture spatiotemporal information simultaneously, presents an effective approach. This mirrors the approach often employed in single-modality video action recognition \cite{korbar2019scsampler} and provides a natural means of fusing information over both space and time. \par

To accommodate heterogeneous modalities, studies have adopted the practice of utilizing separate CNNs for each modality \cite{wang2012mining}. Subsequently, the outputs from these modality-specific networks are combined using fusion layers, often comprising fully connected layers or other fusion techniques. \par

 the incorporation of temporal convolution layers \cite{carreira2017quo} directly models the temporal dynamics within the data. This is critical for recognizing the temporal evolution of actions, a key factor in accurate action recognition. \par

 However, depending on the nature of the deep neural network, data-level, feature-level and decision-level, techniques are often referred to as ``early'' \cite{Feichtenhofer2016twos}, ``middle'' \cite{Roitberg_2019_CVPR_Workshops} or ``intermediate'' \cite{patel2018human}, and ``late'' fusion. Table \ref{tab:datatech} presents some common deep-learning-based MHAR methods that use different data fusion approaches, which are discussed in the following subsections. \par
 
\subsubsection{Early Fusion}
The early-fusion approach captures information and combines it at the feature level \cite{Feichtenhofer2016twos}. In this case, features from different modality sources are concatenated in initial layers into an aggregated feature, which will then be used by the later layers for classification. As this aggregated feature consists of many features, it increases the training and classification time. However, these large aggregates and suitable learning techniques can offer much better recognition performance in the end. In Fig. \ref{fig:dl-fusion}, the initial process of feature fusion is illustrated as an example of early fusion. \par
\subsubsection{Middle Fusion}
The middle-fusion approach merges the features extracted from raw data in the entire neural network so that the higher layers have access to more global information \cite{Roitberg_2019_CVPR_Workshops,patel2018human}. A CNN-based operation is performed to compute the weights and extend the connectivity of all layers. Middle-level fusion attempts to exploit the advantages of early and late fusion in a common framework. 

\subsubsection{Late Fusion}

\begin{figure}
    \centering    \includegraphics[width=0.75\linewidth]{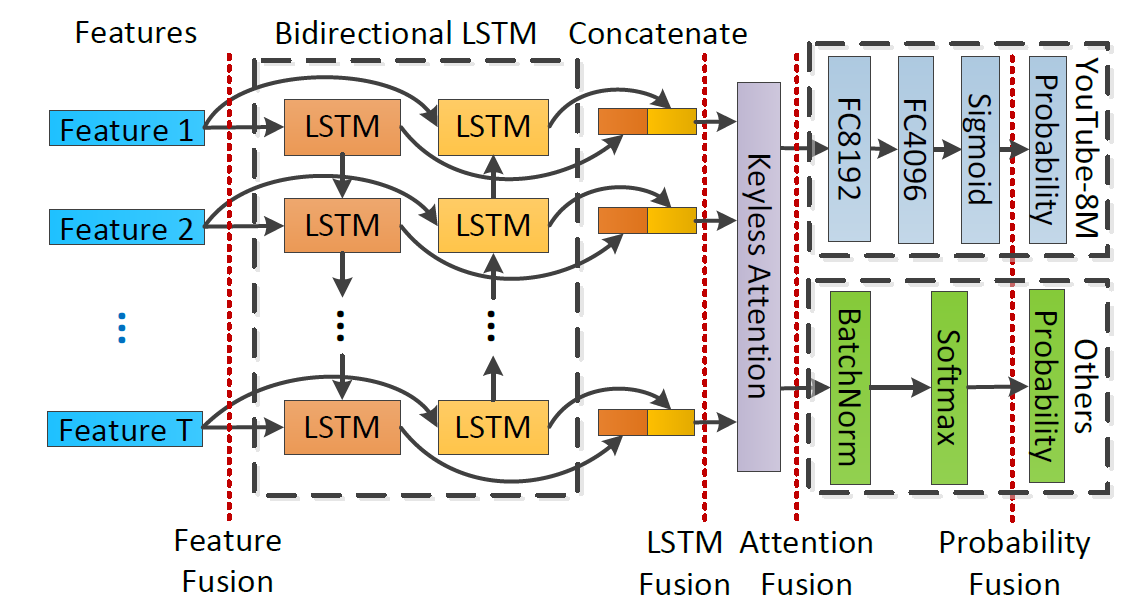}
    \caption{An example of multi-level fusion in deep-learning-based action recognition, where the red dotted lines represent four integration points corresponding to different multimodal fusion methods examined. For a specific
integration point, the network is duplicated for $K$ different modalities, concatenate the features at the integration point, and the network
after the integration point remain unchanged.  (adapted from \cite{LongGMLLLW18})}
    \label{fig:dl-fusion}
\end{figure}

The late-fusion approach indicates a combination of the action information at the deepest layers in the network i.e., after the classification. For example, a typical MHAR network architecture consists of two separate CNN-based networks with shared parameters up to the last convolution layer. The outputs of the last convolution layer of these two separate network streams are processed to the fully connected layer. This step predicts the final output after fusing the classification scores and considers individual class labels at the score layer \cite{shaikh2022maivar}. Different decision rules are deployed to fuse the scores at this stage. Although late fusion ignores some interactions between modality, it adds more simplicity and flexibility in making final decisions when one or more modalities is missing. The late fusion approach has been relatively successful in most MHAR architectures. As evident in Fig. \ref{fig:dl-fusion}, the stage of Probability fusion is an example of late fusion in deep-learning-based methods. \par
% \subsubsection{Hybrid Fusion}

% The hybrid fusion approach is a multi-level fusion approach that fuses the information at different stages throughout the network. In literature, hybrid fusion is often referred to as multi-resolution or multi-level fusion \cite{multi-level-fusion2019}. A recent example of hybrid fusion is shown in Fig. \ref{fig:dl-fusion}. Firstly, features extracted from different modalities are fused, and then the fused features are passed through a bi-directional LSTM module. The output of the LSTM module is again fused and passed to the attention module. After fusing the output from the attention module, the probabilities are fused again at the decision level. The hybrid fusion approach adds relatively more complexity to the network. \par

\subsection{Fusion in Transformers}
% A1
Transformer-based techniques are commonly used for multimodal data fusion. On the one hand, the fusion process typically involves concatenating input sequences from all modalities, or employing a form of cross-attention mechanism (as shown in Fig. \ref{caf}). Attention mechanisms dynamically weigh the importance of features from various modalities, allowing the network to focus on task-relevant information and effectively fuse data. The fusion process can occur at different stages of the architecture, including early, middle, or late stages. In this paper, we discuss both the how and where aspects of the fusion process in Transformer-based multimodal data fusion techniques. \par

\subsubsection{Early Fusion}

This subsection discusses Early Fusion techniques which involve methods in which the fusion occurs prior to input being supplied to the encoder. 

\begin{figure*}[t]
\centering
    \subfloat[\label{ef}]{      \includegraphics[width=0.30\textwidth]{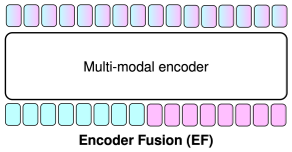}
      }\qquad
    \subfloat[\label{hef}]{      \includegraphics[width=0.30\textwidth]{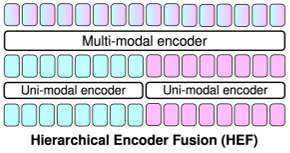}      }\qquad
    \subfloat[\label{caf}]{      \includegraphics[width=0.30\textwidth]{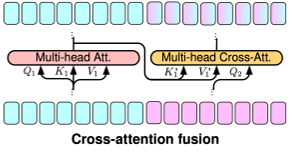}
    }\qquad
    \subfloat[\label{coaf}]{    \includegraphics[width=0.30\textwidth]{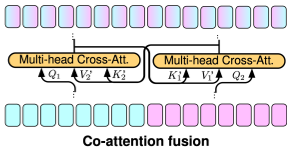}
    }
    \caption{Four main types of performing multimodal fusion in Transformers (adapted from  \cite{selvasurvey2022videotrans}).}
\end{figure*}

\paragraph{Encoder Fusion (EF)} Prior to being fed into the encoder, token embeddings from different modalities are concatenated either in a sequence-wise manner as in \cite{li2020hero,sun2019videobert,liu2021hit} (refer to Fig. \ref{ef}) or in a channel-wise manner as in \cite{fang2019scene}. The former can be likened to the approach of BERT \cite{devlin2018bert} in dealing with pairs of language sentences. To differentiate the tokens belonging to each modality, \cite{sun2019videobert} separator tokens are used, akin to the [SEP] token in BERT, which originally indicates the start of a new sentence but here signals that the succeeding tokens are from another modality. However, most video transformers (VTs) indicate the modality of a given token by summing or concatenating learned modality embeddings in a manner similar to the addition of position embeddings. Encoder fusion significantly increases the computational cost of the self-attention operation up to $O((T_1 + . . . + T_m)^2),$ where $M$ is the number of modalities and $T_m$ is the number of tokens in the $m$-th modality. It is also worth noting here that we not only explore how but also where the fusion is integrated into the architecture, namely at early, middle, or late stages. \par

\paragraph{Hierarchical Encoder Fusion (HEF)} In multimodal learning, Encoder fusion can be performed hierarchically by initially enhancing modality-specific token embeddings on individual encoders, concatenating their outputs, and sending them to a multimodal encoder \cite{lee2020parameter,chen2019contrast,pashevich2021episodic} (see Fig. \ref{hef}). This approach enables intra-modal information to be dealt with before modelling the inter-modal patterns, which is advantageous when handling modalities that are not closely related or accurately aligned at the input level. Experimentation is necessary to determine when to do this. Furthermore, the computational cost is lower than that of the previous encoder fusion by a constant factor, depending on the number of layers in both the unimodal and the multimodal encoders. Although using multiple encoders increases the number of parameters, this can be mitigated by weight sharing \cite{lee2020parameter}.  \par

\subsubsection{Middle Fusion} This subsection discusses Middle Fusion techniques which involve methods in
which the fusion occurs before combining the
outputs of the different encoders.
\paragraph{ Cross-Attention Fusion (CAF)} In the process of modalities fusion using CA, one modality seeks information from another modality to provide context (refer to Fig. \ref{caf}). This simple idea and its flexibility have led to different uses in various studies. In \cite{kim2018multimodal,iashin2020multi}, separate encoder-decoders are proposed for each modality, and each cross-attends to text embeddings before combining the outputs of the different encoders. The study of \cite{jaegle2021perceiver} proposes only one stream that keeps augmenting a small set of latent embeddings by repeatedly cross-attending to the same very long multimodal input sequence of minimal embeddings. In this case, CA layers are interleaved with SA layers that refine the cross-attended information. \cite{zhu2020actbert} proposes a three-stream Transformer, where the central one cross-attends to the other two, and these, in turn, attend to the embeddings generated by their respective opposite previous cross-attentions. In contrast, \cite{camgoz2020multi} also uses three streams, one per modality. However, the fusion is achieved within a master stream that substitutes its SA by asymmetric cross-attention over the other two simultaneously (concatenating both sets of keys and values). \par

\paragraph{Co-Attention Fusion (CoAF)} In contrast to the cross-attention fusion, co-attention involves parallel augmentation of the two modalities by attending to each other's embeddings. In this approach, the CA sub-layer in each stream generates the queries from its own embeddings, whereas keys and values are obtained from the other stream (see Fig. \ref{coaf}). The ViLBERT model \cite{lu2019vilbert} originally proposed this method for images and language, where it has since been adopted by several video-related works \cite{iashin2020better,su2021stvgbert,curto2021dyadformer}. While some works entirely replaced SA with CA \cite{su2021stvgbert,curto2021dyadformer}, others retained it \cite{iashin2020better}. In contrast to encoder fusion, co-attention reduces the computational cost to $O((T_1T_2)^2)$. Additionally, \cite{zhang2021multi} has advocated for co-attending to each other and self-attending to themselves as a means of maintaining intra-modal and inter-modal dynamics separately. \par

\subsubsection{Late Fusion}
All the aforementioned strategies for fusing modalities offer various early and middle fusion options. However, an alternative approach is to perform a late fusion of modalities. This involves running different modalities through parallel encoders and combining their outputs. In classification problems, these outputs could be class score distributions \cite{gavrilyuk2020actor}, as is typically done for Two-stream ConvNets \cite{simonyan2014cnn}. Although this strategy may be suboptimal for Transformers, it may still be useful when there is limited training data available. However, late fusion could be considered by concatenating the augmented aggregation token as proposed by \cite{kim2018multimodal}. \par

There are two concatenation-based approaches, namely Encoder Fusion (EF) and Hierarchical Encoder Fusion (HEF), the former of which concatenates the modalities at an early stage while the latter concatenates them at a middle stage, which is more computationally intensive but eliminates the need to experimentally determine the fusion point. In contrast, CAF and CoAF are based on cross-attention mechanisms and are inherently limited to fusing only two modalities. Nonetheless, recent studies have attempted to extend these methods to more than two modalities, as evidenced by the cumbersome approach proposed by \cite{zhu2020actbert} in their ActBERT model. Compared to EF and HEF, both CAF and CoAF are more efficient, but they still suffer from the issue of identifying the optimal fusion point. However, middle fusion may be useful when the input modalities are asynchronous or differ in nature.

%% file: 5.challenge.tex
\section{Challenges, Future Directions And Limitations}
\label{five}

In this section of the survey article, we will delve into the potential challenges, future directions associated with the field under discussion and limitations of this survey. By exploring the challenges, we aim to shed light on the hurdles that must be overcome to make further progress. Further, by discussing future directions, we aim to provide insights into the exciting new avenues for research and development that lie ahead. Finally, by acknowledging the limitations of this survey, we aim to provide a balanced and realistic view of its current scope and limitations.

% A3
\subsection{Challenges and Future Directions}

\vspace{0.7mm}
\noindent{\bf Future of Datasets:}  Extensive and comprehensive datasets hold significant importance for the advancement of MHAR, particularly for MHAR methods based on deep learning. Several factors, such as the  magnitude, diversity, applicability, and modality type, indicate its quality. Despite the availability of numerous datasets that have significantly progressed the MHAR field, the development of novel benchmark datasets is still necessary to further promote research in this domain. For instance, most existing multimodal datasets have been obtained in controlled settings with voluntary participants performing actions. Consequently, gathering multimodal data from uncontrolled environments to establish substantial and challenging benchmarks for enhancing MHAR's practical applications would be beneficial. Moreover, constructing large and complex datasets for recognizing each person's actions in crowded settings warrants further exploration. \par

\vspace{0.7mm}
\noindent{\bf Multimodal Learning:} Earlier discussions have outlined various multimodal learning methods, including multimodal fusion and cross-modality transfer learning, proposed for MHAR. Multimodal data fusion, which can complement each other, has been shown to enhance MHAR performance while co-learning can address data deficiencies in certain modalities. Nevertheless, as \cite{wang2020makes} has highlighted, several challenges impede the effectiveness of many existing multimodal methods, such as the risk of overfitting. This underscores the need to develop more effective strategies for multimodal fusion and co-learning in MHAR. This challenge requires further exploration and innovation to develop advanced techniques for addressing the limitations of the existing state-of-the-art multimodal methods.

\vspace{0.7mm}

\vspace{0.7mm}
\noindent{\bf Labelled Data Scarcity in Unsupervised and Semi-Supervised Learning:} The application of supervised learning methods, particularly deep learning-based ones, typically necessitates a substantial amount of labelled data for model training, which can be expensive to obtain. Conversely, unsupervised and semi-supervised learning techniques \cite{alayrac2016unsupervised,singh2021semi,song2021spatio} have the potential to utilize unlabelled data to train models, thereby mitigating the requirement for extensive labelled datasets. Since unlabelled action samples are often more easily collected than labelled ones, unsupervised and semi-supervised MHAR has emerged as a crucial research direction that merits further investigation. 
% A4
% \vspace{0.7mm}
% \noindent{\bf Interpretability :} Exploring the attention mechanism of Transformers enables researchers to gain insights into the important aspects of the input that the model deems relevant, even though this does not necessarily provide a deep understanding of the learned relationships \cite{jain2019attention}. While some works have attempted to interpret Transformers for vision, the literature on Vision Transformers (VTs) offers only a limited number of studies that visualize attention activations for specific samples \cite{pashevich2021episodic}. Notably, the work of \cite{li2020hero} has revealed six distinct patterns in cross-modal attention, indicating how certain heads learn local, modality-specific, or cross-modal attention. Other studies explored the contributions of multiple experts to the final representation \cite{gabeur2020multi}, how cropping augmentations can help the model learn audio sources tied to specific spatiotemporal regions of the video \cite{patrick2021space}, and the use of rollout to aggregate attention over multiple heads \cite{girdhar2021anticipative}. Nevertheless, the issue of interpretability of Transformers, including VTs, remains open \cite{bracsoveanu2020visualizing}.

\vspace{0.7mm}
\noindent{\bf Generalization:} It is acknowledged that while some works have investigated the generalization capabilities of Transformers, this issue remains open and understudied, particularly in the case of VTs. Although some studies have examined the generalization of Transformers in natural language processing, such as out-of-distribution (OOD) generalization \cite{hendrycks2020pretrained} and cross-modal transfer learning with minimal fine-tuning \cite{lu2021pretrained}, the analysis of the generalization capabilities of VTs is scarce. To the best of our knowledge, only one work has thoroughly investigated the topic of generalization in VTs \cite{zhang2022delving}.

\vspace{0.7mm}
\noindent{\bf Self-Supervised Approaches:} There is a need for further investigation about video self-supervised tasks and their applicability to VTs. While contrastive techniques dominate self-supervised approaches, there are several untapped possibilities for utilizing self-supervised video tasks on VTs, including temporal consistency \cite{fernando2017self}, interframe predictability \cite{han2019video}, geometric transformations \cite{kim2019self} and motion statistics \cite{wang2019self}. Additionally, the recent emergence of self-supervised vision tasks like SimCLR \cite{chen2020simple} and Barlow Twins \cite{zbontar2021barlow} for images, or BraVE \cite{recasens2021broaden} for video, warrants further exploration for their application to VTs.

\vspace{0.7mm}
\noindent{\bf Online and Mobile MHAR:} While some studies have explored the use of deep MHAR on mobile devices, such as smartphones and watches, they still face significant limitations in online and mobile deployment \cite{lane2015deepear, bhattacharya2016smart}. In these approaches, the model is trained offline on a remote server, and the mobile device only employs the trained model, which is neither real-time nor efficient for incremental learning. Two potential solutions to address this challenge are to reduce the communication cost between mobile and server and to improve the computing ability of mobile devices. \par

\vspace{0.7mm}
\noindent{\bf Collaboration in Deep-Shallow MHAR:} High computational requirements of deep models are often a bottleneck, rendering them unsuitable for wearable devices. Contrastingly, shallow neural networks (NN) and traditional pattern recognition (PR) methods are not able to achieve high-performance levels. Therefore, a middle ground needs to be found by developing lightweight deep models that can still achieve high-performance levels while being efficient enough to run on mobile devices. In this regard, the collaborative efforts of deep and shallow models have the potential to provide accurate and lightweight MHAR solutions. However, several issues need to be addressed, such as how to effectively share parameters between deep and shallow models. Moreover, the existing offline training of deep models hinders real-time execution, necessitating the need for online training approaches that can enhance the adaptability of the model to new environments. \par

\vspace{0.7mm}
\noindent{\bf Effective Transfer Learning:} The process of data annotation via transfer learning is executed through the utilization of labelled data from auxiliary domains \cite{wang2017balanced}. Various factors related to human activity can be leveraged as supplementary information through deep transfer learning. This approach poses several challenges, including the sharing of weights between networks, the effective exploitation of knowledge between activity-related domains, and the identification of relevant domains. These issues require resolution in order to facilitate the effective application of transfer learning methodologies. \par

\vspace{0.7mm}
\noindent{\bf Hybrid Sensor and  Context-Aware MHAR:} The comprehensive information obtained from hybrid sensors is of considerable utility in discerning fine-grained activities \cite{vepakomma2015wristocracy}. Particular emphasis must be placed on the recognition of such activities through the collaborative utilization of hybrid sensors. In contrast, context refers to any information that can be employed to describe the circumstances surrounding an entity \cite{abowd1999towards}. Contextual data sources, such as Wi-Fi, Bluetooth, and GPS, can provide valuable insight into environmental factors associated with a given activity. The effective exploitation of contextual information can greatly enhance the ability to recognize both user states and specific activities. \par

\vspace{0.7mm}
\noindent{\bf Beyond Activity Recognition:}
The recognition of activities frequently represents the initial stage in a variety of applications. For example, professional skill assessment is necessary for fitness training, while smart home assistants are indispensable components of healthcare services. Early efforts in activity recognition include research on climbing assessment \cite{khan2015beyond}. Recent investigations suggest that leveraging the expertise of crowds can substantially facilitate the task \cite{prelec2017solution}. Crowdsourcing offers a means of annotating unlabelled activities by utilizing the collective abilities of a large group. In addition to passive label acquisition, researchers may also develop more sophisticated and privacy-preserving methodologies to collect valuable labels. As deep learning continues to advance, activity recognition applications can be expanded beyond simple recognition tasks. \par 

\subsection{Limitations}

Our study was limited to research papers published within the last 10 years (2012-2022) and only included papers in English, potentially excluding some studies in other languages. Only studies that utilized visual data were considered and others, such as those using olfactory, infrared, or tactile data, were outside the scope of this research. A potential limitation is publication bias, which may lead to an overestimation of the advantages of using multiple forms of data for analysis. The studies reviewed employed different input methods and evaluated various methods for recognizing actions on different datasets, making direct comparison of results difficult. Additionally, not all articles provided statistical confidence intervals, making it challenging to compare their findings.

%% file: 6.conclusion.tex
\section{Conclusion}
\label{six}

% summary of findings
% limitations and possible improvements
% Limitations: 
% Broader Impact

In this survey, we have conducted a comprehensive analysis and synthesis of the primary advancements and emerging trends in adapting Convolutional Neural Networks (CNNs) and Transformers for multimodal recognition of human actions. Drawing upon the existing literature, we have devised a robust taxonomy for CNN and Transformer architectures based on their modality, intended task and overall structure. Furthermore, we have investigated diverse techniques for embedding, encoding and fusing different multimodal representations to achieve more accurate recognition of human actions. \par
In addition, we have provided a comparative analysis of the leading approaches on different datasets and suggested key design modifications necessary for improving their performance. Moreover, we have discussed the current trends and potential challenges associated with the different components of the multimodal action recognition pipeline. Despite the considerable progress that has been made, the potential of multimodal fusion for action recognition remains largely unexplored, and there remain several significant challenges that need to be addressed. Finally, we express a keen interest in comparing Transformer architectures to CNNs for multimodal understanding, given the tremendous promise of global-based learning methods.